\documentclass[letterpaper]{article} 
\usepackage{aaai24}  
\usepackage{times}  
\usepackage{helvet}  
\usepackage{courier}  
\usepackage[hyphens]{url}  
\usepackage{graphicx} 
\usepackage{natbib}  
\usepackage{caption} 
\usepackage{amstext}
\usepackage{varwidth}
\usepackage{tcolorbox}
\usepackage{multirow}

\newcommand{\datasetFont}{\text}
\newcommand{\ours}{\datasetFont{PromptGAT}\xspace}
\newcommand{\vanilla}{\datasetFont{VanillaGAT}\xspace}

\newcommand{\metricsFont}{\textit}
\newcommand{\att}{\metricsFont{ATT}\xspace}
\newcommand{\reward}{\metricsFont{Reward}\xspace}
\newcommand{\tp}{\metricsFont{TP}\xspace}
\newcommand{\queue}{\metricsFont{Queue}\xspace}
\newcommand{\delay}{\metricsFont{Delay}\xspace}

\usepackage{microtype}
\usepackage{subfigure}
\usepackage{booktabs} 
\usepackage{graphicx}
\usepackage{booktabs}
\usepackage{amsmath}
\usepackage{amssymb}
\usepackage{color}
\usepackage{longtable}
\usepackage{etoolbox}

\usepackage[ruled,linesnumbered,lined]{algorithm2e}

\newcommand{\weather}[1]{\textit{\textcolor{blue}{#1}}}
\newcommand{\road}[1]{\textit{\textcolor{purple}{#1}}}
\newcommand{\traffic}[1]{\textit{\textcolor{red}{#1}}}

\title{Prompt to Transfer: Sim-to-Real Transfer for Traffic Signal Control with Prompt
Learning}

\author{
    Longchao Da \textsuperscript{\rm 1},
    Minquan Gao \textsuperscript{\rm 2},
    Hao Mei \textsuperscript{\rm 1},
    $\text{Hua Wei \textsuperscript{\rm 1}}^{\text{\textdagger}}$
}
\affiliations{
    \textsuperscript{\rm 1}Arizona State University, 350 E Lemon St, Tempe, AZ 85287, U.S.\\
    \textsuperscript{\rm 2}Johns Hopkins University, 3400 N. Charles Street
Baltimore, MD 21218, U.S.\\
    
   ${}^{\text{\textdagger}}$Corresponding email: hua.wei@asu.edu\\
}

\begin{document}

\maketitle

\begin{abstract}
Numerous solutions are proposed for the Traffic Signal Control (TSC) tasks aiming to provide efficient transportation and mitigate congestion waste. Recently, promising results have been attained by Reinforcement Learning (RL) methods through trial and error in simulators, bringing confidence in solving traffic congestion in cities. However, there still exists a performance gap when simulator-trained policies are deployed to the real world. This issue is mainly introduced by the system dynamic difference between the training simulator and the real-world environments. The Large Language Models (LLMs) are trained on mass knowledge and proved to be equipped with astonishing inference abilities.
In this work, we leverage LLMs to understand and profile the system dynamics by a prompt-based grounded action transformation. Accepting the cloze prompt template, and then filling in the answer based on accessible context, the pre-trained LLM's inference ability is exploited and applied to understand how weather conditions, traffic states, and road types influence traffic dynamics, being aware of this, the policies' action is taken and grounded based on realistic dynamics, thus help the agent learn a more realistic policy. We conduct experiments using DQN to show the effectiveness of the proposed PromptGAT's ability to mitigate the performance gap from simulation to reality (sim-to-real).

\end{abstract}

\section{Introduction}
Traffic Signal Control (TSC) is a critical task aimed at improving transportation efficiency and alleviating congestion in urban areas~\cite{zhang2020generalight}. Reinforcement Learning (RL) methods have shown promising results in tackling TSC challenges through trial and error in simulators~\cite{ghanadbashi2022using, mei2022libsignal, noaeen2022reinforcement, ducrocq2023deep, zang2020metalight,wu2020multi, haydari2020deep}, bringing hope for solving cities' traffic congestion issues. Simulation is a valuable tool for control tasks in the real world because the execution of a control skill in simulation is comparatively easier than real-world execution. However, a notable performance gap arises when deploying simulator-trained policies to real-world environments~\cite{da2023uncertainty}, mainly due to differences in system dynamics between training simulators and the actual road conditions.

Grounded Action Transformation (GAT) is a framework designed to address the performance gap that arises when transferring policies learned in simulation to real-world scenarios (sim-to-real)~\cite{hanna2017grounded}. The key idea behind GAT is to induce simulator dynamics to resemble those of the real world, where policy learning takes place in simulation, and dynamics learning relies on real-world data.

\begin{figure}[t!]
    \centering
    \includegraphics[width=0.48\textwidth]{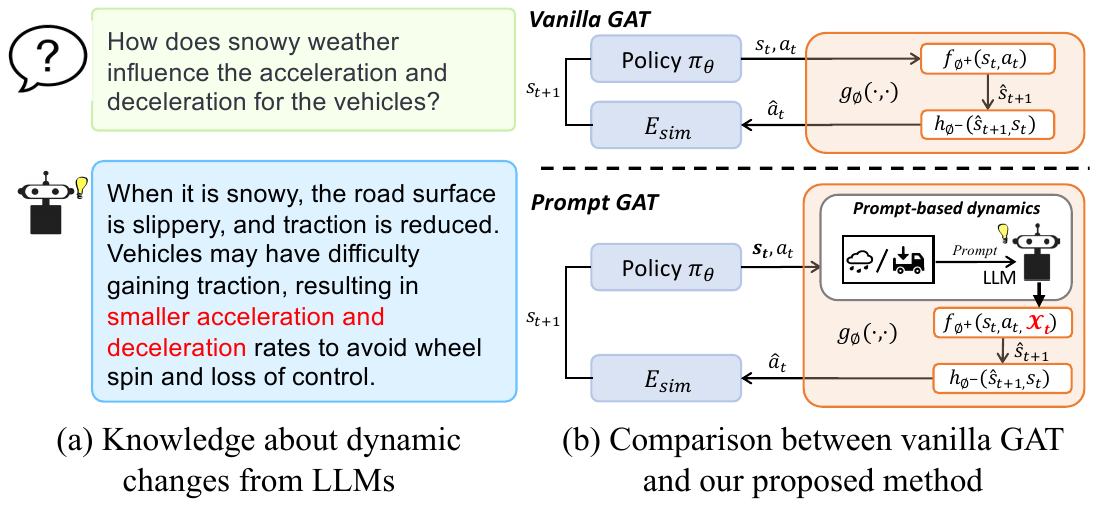}
    \vspace{-5mm}
    \caption{Integrating knowledge from LLMs into GAT. (a) LLMs have implicit human knowledge about the change in dynamics. (b) Comparisons between vanilla GAT and our proposed PromptGAT, with GAT integrating a prompt-based dynamics modeling module.}
\label{fig:intro}
\end{figure}

The dynamics model, also known as the forward model $f_{\phi^+}$, plays a crucial role in the GAT framework. It takes the current state $s_t$ and action $a_t$ as inputs and predicts the possible next state $s_{t+1}$ in the real world.
Traditional GAT methods focus on learning $f_{\phi^+}$ solely based on real-world data. While these approaches enable the forward model to be accurately fitted to real-world dynamics, it requires a substantial amount of real-world data covering the entire state distribution to achieve accurate predictions.

One limitation of conventional GAT methods is their struggle to handle unobserved states in the real world. When the policy encounters states that have not been previously observed in the real-world data, the learned forward model may predict $s_{t+1}$ with significant errors. This is particularly evident under extreme weather conditions or rare events that are infrequently represented in the training data.
In contrast, \emph{human knowledge} allows us to infer the behavior of the system under such unique conditions. For example, we as humans understand that during extreme weather, vehicles tend to move slower with smaller acceleration and deceleration rates, and the same duration of a green traffic signal may result in a smaller throughput. Moreover, humans can reason that adjusting the duration of green lights from the policy might be necessary to achieve a similar performance as observed in the simulation.

To leverage this implicit human knowledge for more accurate forward models, we propose a prompt-based GAT method, known as PromptGAT by introducing Large Language Models (LLMs) into the GAT framework. Specifically, as is shown in Figure~\ref{fig:intro}(b), in the learning of the forward model, we design a prompt-based dynamics modeling module to better understand the real-world dynamic by asking LLMs how weather conditions, traffic states, and road types influence traffic dynamics. Through the inference ability of LLMs in profiling the system dynamics, the agent is able to learn the grounded action in GAT based on a more accurate and general forward model. This process facilitates the learning of more realistic policies and enhances the transferability of RL models from simulation to reality.
In summary, the contribution of this paper is as follows:
\\\noindent$\bullet$~This paper proposes a novel method, \ours, to mitigate the sim-to-real transfer problem in the context of traffic signal control by incorporating human knowledge with LLMs. To the best of our knowledge, this is the first paper bridging the performance gap between simulation and real-world settings in traffic signal control with LLMs.
\\\noindent$\bullet$~This paper provides the design of prompt generation and dynamics modeling module to understand the change of dynamics in the sim-to-real transfer. Leveraging LLMs with prompt and chain-of-thought~\cite{wei2022chain}, \ours provides valuable insights into the system dynamics, which enhances the agent's understanding of real-world scenarios.
\\\noindent$\bullet$~We conduct extensive experiments and case studies to validate the performance of our approach and showcase its potential impact on traffic signal control. All the experiments are conducted under a simulation-to-simulation setting with reproducible experiment settings. All the data and code can be found in repository~\footnote{\url{https://github.com/DaRL-LibSignal/PromptGAT}}. 

\section{Related Work}

This section will introduce the related work from three aspects, regarding the traffic signal control methods, simulation-transfer methods, and prompt learning techniques. 

\paragraph{Traffic Signal Control Methods} 
Optimizing traffic signal policies to mitigate traffic congestion has posed a significant challenge within the realm of transportation. Diverse methodologies have been thoroughly investigated, encompassing rule-based methods~\cite{dion2002rule, chen2020toward} as well as RL-based methods~\cite{wei2019colight,wei2019presslight,wei2018intellilight} for enhancing vehicle travel time or reducing delays, most of which yielded notable enhancements over pre-existing time control techniques. 
Although most of the current RL-based TSC methods do not consider the sim-to-real gap problem, a few recent studies start to tackle the sim-to-real gap by modifying the simulator directly~\cite{muller2021towards, mei2022libsignal}, requiring the parameters of the simulator can be easily modified to perfectly match the real world. Rather than modify the simulator, this paper proposes to modify the output actions of the policies learned in the traffic simulator.

\paragraph{Sim-to-real Transfer}
Mainly three categorized groups of literature exist in the sim-to-real transfer domain~\cite{zhao2020sim}. The first group is \textbf{\emph{domain randomization}}~\cite{tobin2019real, andrychowicz2020learning}, with the objective of training policies capable of adapting to environmental variations. This strategy primarily relies on simulated data and proves advantageous when dealing with uncertain or evolving target domains. The second group is \textbf{\emph{domain adaptation}}~\cite{tzeng2019deep, han2019learning}, which is dedicated to addressing the challenge of domain distribution shift by aligning features between the source and target domains. Many domain adaptation techniques focus on narrowing the gap in robotic perception~\cite{tzeng2015towards,fang2018multi,bousmalis2018using,james2019sim}, in TSC, the gap mainly attributed to the dynamics rather than perception since most TSC methods directly regard the vectorized representations like lane-level number of vehicles or delays as observations. Recently, \textbf{\emph{grounding methods}}, known as the third group of approaches, improves the accuracy of the simulator concerning the real world by correcting for simulator bias. Unlike system identification approaches~\cite{6907423, Cully_2015} that try to learn the accurate physical parameters, Grounded Action Transformation (GAT)~\cite{hanna2017grounded} induces the dynamics of the simulator to match the real world with grounded action rather than require a parameterized simulator that can be modified. It has demonstrated promising results for sim-to-real transfer in robotics.  Our \ours is inspired by GAT, with novel designs on leveraging LLMs to enhance action transformation.

\paragraph{Prompt Learning}
Prompt learning, first introduced by~\cite{petroni-etal-2019-language}, has been widely studied in NLP during these years~\cite{jiang2020can, shin-etal-2020-autoprompt}. Prompting means prepending instructions to the input and pre-training the language model so that the downstream tasks can be promoted. ~\cite{poerner2019bert} use manually defined prompts to improve the performance of language models. To adapt LLMs for specific applications, developers often send prompts (aka, queries) to the model, which can be appended with domain-specific examples for obtaining higher-quality answers. A collection of prompt management tools, such as ChatGPT Plugin \cite{ChatGPTplugins}, GPT function API call \cite{apiCalls}, LangChain \cite{langchain}, AutoGPT \cite{Auto-GPT}, and BabyAGI \cite{BabyAGI}, have been designed to help engineers integrate LLMs in applications and services. As far as we know, there is no exploration for prompt learning in sim-to-real transfer or traffic signal control task. 


\begin{figure*}[htbp]
    \centering
    \includegraphics[width=0.95\textwidth]{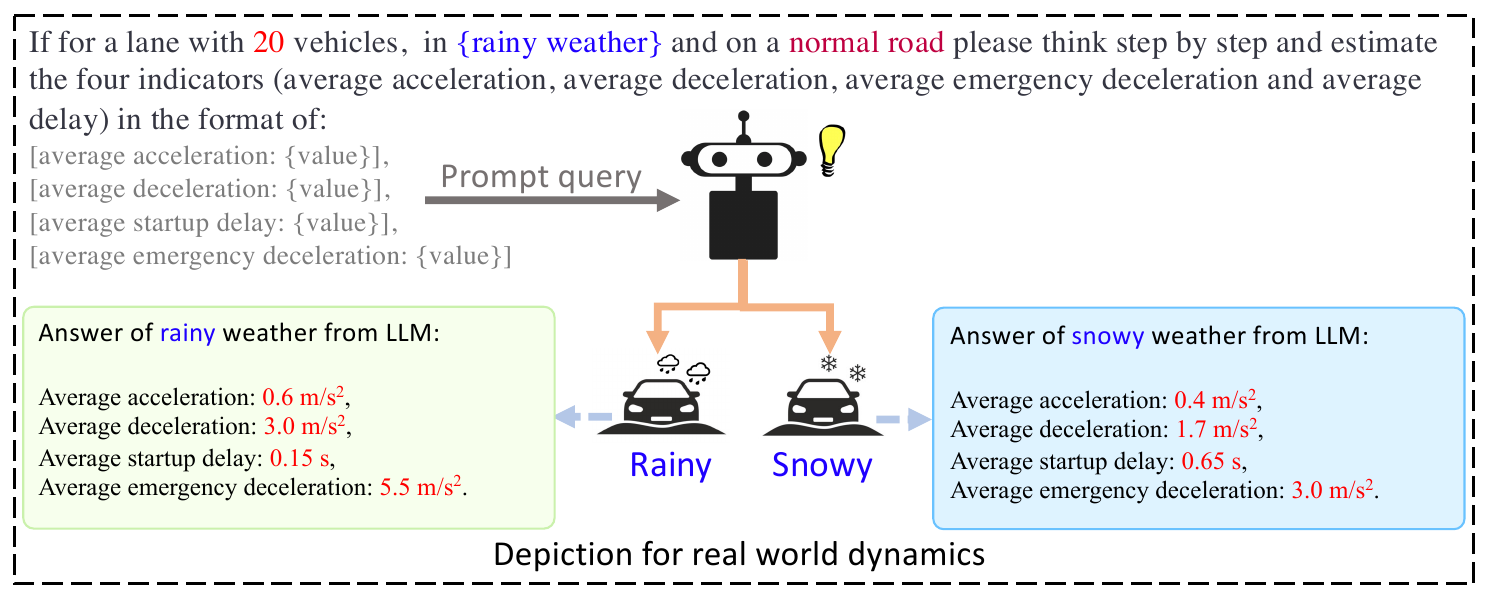}
    \caption{An example of using LLM with a prompt template for answers to depict real-world dynamics by providing traffic state (vehicle number), weather type, and road type to induce the LLM to infer based on domain knowledge. Given the same vehicle quantity and road type, we could observe that the answers under different weathers abide by the reality situation that snowy weather is more severe than rainy weather.}
    \label{fig:enter-label}
\end{figure*}

\section{Method}

\subsection{Preliminaries}

\subsubsection{RL-based Traffic Signal Control} In Traffic Signal Control (TSC), controllers determine intersection phases. Each phase comprises predefined, non-conflicting traffic movement combinations. The signal controller tends to adopt a phase that minimizes the average queue length in a certain traffic range during the next time interval $\Delta t$. The literature~\cite{wei2018intellilight,chen2020toward,zheng2019learning,wei2019colight}, propose to distribute each traffic signal with an agent, responsible for choosing the actions (phases) in each time frame. It is a common practice to characterize the TSC problem in a MDP by $\mathcal{M}  = \langle \mathcal{S}, \mathcal{A}, P, r, \gamma \rangle$ where $\mathcal{S}$ denotes the system state space $\mathcal{S}$, $\mathcal{A}$ denotes the set of action space,  $P$ denotes as the transition dynamics describing the probability distribution of next state $s_{t+1} \in \mathcal{S}$, $r$ denotes the reward, and $\pi_{\theta}$ as the policy parameterized by $\theta$ and $\gamma$ is the discount factor.

An RL algorithm expects to learn an optimal policy that maximizes the long-term expectation of accumulated reward which is adjusted by a discount factor $\gamma$. The discounted accumulated reward is $\mathbb{E}_{(s_t, a_t)\sim (\pi_{\theta},\mathcal{M})}[ \sum_{t=0}^T \gamma^{T-t} r_t(s_t, a_t) ]$. We follow the past work which defines $\mathcal{A}$ as discrete action spaces, and use Deep Q-network (DQN)~\cite{wei2018intellilight} to optimize the RL policy. The above procedure is conducted in the simulation environment $E_{sim}$ in previous work.

\subsubsection{Grounded Action Transformation}
The Grounded Action Transformation (GAT) framework, initially introduced in the field of robotics, aims to enhance robotic learning through the utilization of real-world $E_{real}$ trajectories to adapt and modify the simulated environment $E_{sim}$. Within the GAT framework, the Markov Decision Process (MDP) within $E_{sim}$ is considered imperfect yet adjustable, allowing for its parameterization as a transition dynamic $P_{\phi}(\cdot|s, a)$. Based on the real-world dataset $\mathcal{D}_{real} =  \{\tau^{1}, \tau^{2}, \dots, \tau^{I}\}$, where $\tau^{i} = (s_0^{i}, a_0^{i}, s_1^{i}, a_1^{i}, \dots, s_{T-1}^{i}, a_{T-1}^{i}, s_T^{i})$ is a trajectory collected by executing policy $\pi_{\theta}$ in $E_{real}$, GAT intends to find $\phi^*$ that minimize differences between transition dynamics:

\begin{equation}
\phi^* = \arg \min_{\phi} \sum_{\tau^i \in \mathcal{D}_{real}} \sum_{t=0}^{T-1} d(P^*(s^i_{t+1}|s^i_t, a^i_t), P_{\phi}(s^i_{t+1}|s^i_t, a^i_t))
\end{equation}

\noindent where $d(\cdot)$ represents the two dynamics' discrepancy, $P_{\phi}$ stands for the simulation transition dynamics, and $P^*$ stands for the real world transition dynamics. 


For efficient determination of $\phi$, the GAT takes the agent's current state $s_t$ and action $a_t$ predicted by the policy $\pi_\theta$ as input. It then produces a grounded action, denoted as $\hat{a}_t$, through the use of an action transformation function that is parameterized by $\phi$:

\begin{equation}
\label{eq:gat}
    \hat{a}_t = g_{\phi}(s_t, a_t) = h_{\phi^{-}}(s_t, f_{\phi^{+}}(s_t, a_t))
\end{equation}
 which includes two specific functions: a forward model $f_{\phi^{+}}$, and an inverse model $h_{\phi^{-}}$, as is shown in Figure~\ref{fig:intro}.   
 
\noindent $\bullet$ \textit{The forward model} $f_{\phi^{+}}$ is trained with the data from $E_{real}$, with the purpose of predicting the next possible state $\hat{s}_{t+1}$ given current state $s_t$ and action $a_t$:

\begin{equation}
\label{eq:forward}
     \hat{s}_{t+1} =  f_{\phi^{+}}(s_t, a_t)
\end{equation}

\noindent $\bullet$ \textit{The inverse model} $h_{\phi^{-}}$ is trained with the data from $E_{sim}$, with the purpose of predicting the action $\hat{a}_t$ that brings about the given next state from the current state $s_t$. Specifically, the inverse model in GAT takes $\hat{s}_{t+1}$, the output from the forward model, as its input for the next state: 
\begin{equation}
\label{eq:inverse}
     \hat{a}_t =  h_{\phi^{-}}(\hat{s}_{t+1}, s_t)
\end{equation}

\begin{figure*}[t!]
    \centering
    \includegraphics[width=0.95\textwidth]{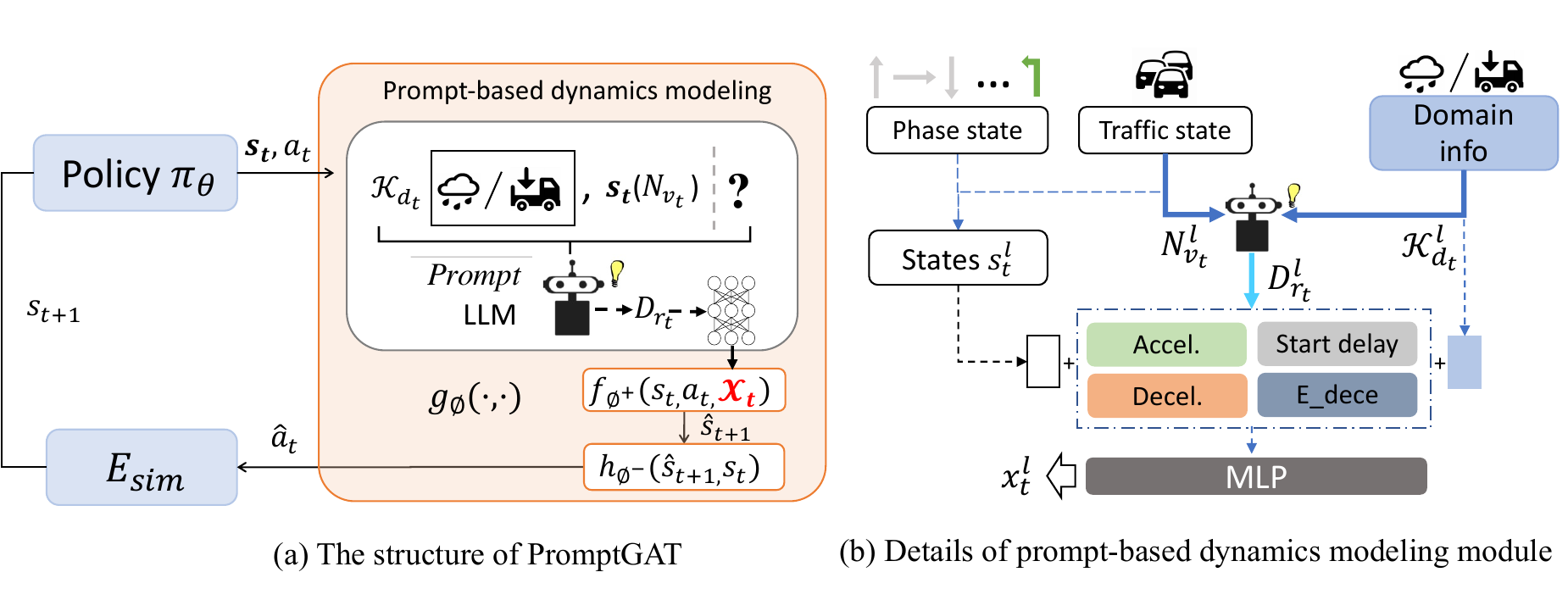}
    \vspace{-5mm}
    \caption{The overall framework of our proposed \ours. (a) The structure of \ours, with a prompt0based dynamics modeling integrating the knowledge of LLMs into the learning of forward model $f_{\phi^+}$. (b) Details of prompt-based dynamics modeling module that infer and integrate the change of dynamics with traffic states.}
    \vspace{-5mm}
    \label{fig:fuison}
\end{figure*}
Based on the state action pair $<s_t, a_t>$ where the $a_t$ is predicted by the policy $\pi_\theta$, the grounded $\hat{a}_t$  takes place in $E_{sim}$ will rectify the $s_{t+1}$ in $E_{sim}$ close to the predicted next state $\hat{s}_{t+1}$ in $E_{real}$, making the dynamics $P_{\phi}(s_{t+1}|s_t, \hat{a}_t)$ in simulation close to the real-world dynamics $P^*(\hat{s}_{t+1}|s_t, a_t)$. By this, the policy $\pi_\theta$ learned in $E_{sim}$ with $P_{\phi}$ is close to $P^*$, and will have a smaller performance gap when transferred to $E_{real}$ with $P^*$.

\subsection{Prompt-based GAT}

\subsubsection{In-context Learning for Dynamics Knowledge}
LLMs are known to be capable of in-context zero-shot or few-shot abilities, which adapt these models to diverse tasks without gradient-based parameter updates~\cite{alayrac2022flamingo}. This allows them to rapidly generalize to unseen tasks and even exhibit apparent reasoning abilities with appropriate prompting strategies. Here is a set of system dynamic descriptions in the natural language as in Equation~\eqref{eq:promptpair}, provided to GPT-4.0 as context. They respectively stand for the scene weather information, road condition, and lane-level traffic state: \begin{equation}\label{eq:promptpair}
  \textit{Context}: \langle \weather{Weather}\rangle \langle \road{Road Type} \rangle \langle \traffic{Traffic State}\rangle 
\end{equation}
These contexts are organized and filled into the designed prompt template as in below: 
\begin{equation}\label{eq:define}
    \langle \textit{Task}\rangle \langle[\textit{Context}]\rangle \langle \textit{Output Restriction}\rangle,
\end{equation}
where $ \langle \textit{Task}\rangle$ provides task intention explanation to LLMs, and $\langle \textit{Output Restriction}\rangle$ induces the LLMs to infer the possible dynamics change based on the $\langle[\textit{Context}]\rangle$ provided in Equation~\eqref{eq:promptpair} for the language model to understand the current perceptible information. As shown in Figure~\ref{fig:enter-label}, the resulting output dynamics knowledge could be then utilized by the forward model $f_{\phi^+}$. 


\subsubsection{Prompt-based Dynamics Modeling}
In GAT, the learning of the forward model $f_{\phi^+}$ and inverse model $h_{\phi^-}$ are crucial. 
The forward model $f_{\phi^+}(s_t, a_t)$ in GAT predicts the next RL state $\hat{s}_{t+1}$ in the real world given taken action $a_t$ and the current state $s_t$ as in Equation~\eqref{eq:forward}. In the above method, direct prediction only using $(s_t, a_t)$ omits the consideration of domain knowledge $\mathcal{K}_d$, such as weather or road conditions, but the state transition $s_{t+1} = T_r(s_t, a_t)$  in the real world is a joint consequence related to this perceptible domain knowledge and real-time traffic states (vehicle quantities), e.g., In the snowy days, vehicles normally act with larger startup delay than in good weather time, and for vehicles on the high occupied road, the acceleration and emergency deceleration will be lower than those on the free road that's mainly decided by the vehicles' standard machine characteristics. Therefore, we propose leveraging the $\mathcal{K}_d$ to provide a hint on the concrete real-world system dynamics $D_r$ such as \textit{acceleration}, \textit{deceleration}, \textit{emergency deceleration} and \textit{startup delay} reflected by transition $T_r$. For $ {\forall}$ $ s_t \in S$ on lane $l$, we employ LLM (GPT-4.0) to realize inference by the prompt organized in Equation~\eqref{eq:define}: 
\begin{equation}\label{eq:llm}
    D_{r_t}^l = LLM(Prompt(\mathcal{K}_{d_t}^l, N_{v_t}^l))
\end{equation}
where $\mathcal{K}_{d_t}^l = \textit{(weather}, \textit{road)}$ and $N_{v_t}^l$ is the number of vehicles. 
Based on this, we incorporate the current lane state, lane level the domain knowledge $\mathcal{K}_{d_t}^l$ and dynamics knowledge $D_{r_t}^l$ from LLM together into the forward model through a fusion module of the model's network design as in Figure~\ref{fig:fuison}. Please note that the \textit{road} description in $\mathcal{K}_{d_t}^l$ can vary for lanes but keep the same for the whole complete trajectory, and \textit{weather} holds on same for one complete trajectory as well.
\begin{equation}\label{eq:fuison1}
    \dot{x_t}^l = \oplus \{s_t^l, D_{r_t}^l, \mathcal{K}_{d_t}^l\}
\end{equation}
\begin{equation}\label{eq:fuison2}
    {x_t}^l = ReLU(Linear(\dot{x_t}^l))
\end{equation}
where $\dot{x}$ is a temporary calculation step after the operation $\oplus$, which is implemented as $Concatenate(\cdot)$, the derived ${x_t}^l$ represents the feature space for specific lane $l$ at time step $t$. The input for $f_{\phi^+}$ is an integrated information $\mathcal{X}$ from all $n$ lanes: $\mathcal{X}_t = (x_t^1, x_t^2, \ldots, x_t^n)$. Now we could represent the  forward model into:
\begin{equation}
\label{eq:forward2}
     \hat{s}_{t+1} =  f_{\phi^{+}}(s_t, a_t, \mathcal{X}_t)
\end{equation}

The $f_{\phi^+}$ is approximated by a NN (neural network) and  $\phi^+$ is optimized through minimizing the Mean Squared Error (MSE) loss, in the equation, the $s^i_t$, $a^i_t$, $s^i_{t+1}$ are sampled from the trajectories collected from $E_{real}$:
\begin{equation}
\label{eq:forward-loss}
     \mathcal{L}(\phi^+) = MSE(\hat{s}^i_{t+1}, s^i_{t+1}) = MSE(f_{\phi^+}(s^i_t, a^i_t), s^i_{t+1})
\end{equation}

Different from $f_{\phi^+}$, the inverse model $h_{\phi^-}(\hat{s}_{t+1}, s_t)$ predicts the grounded action $\hat{a}^i_t$ that can lead to the same traffic states $\hat{s}_{t+1}$ in simulation $E_{sim}$. $h_{\phi^-}$ could be learned through the interactions within simulation with lower cost than $f_{\phi^+}$, therefore in this paper, we did not incorporate the dynamics knowledge in the inverse model for a more accurate model. The $h_{\phi^-}$ is also approximated by NN and $\phi^-$ is optimized by minimizing the Categorical Cross-Entropy (CE) loss since the target $a^i_t$ is a discrete value in traffic signal control problem defined by existing work~\cite{wei2018intellilight}:
\begin{equation}
\label{eq:inverse-loss}
    \mathcal{L}(\phi^-) = CE(\hat{a}^i_t, a^i_t) = CE(h_{\phi^-}(s^i_{t+1}, s^i_t), a^i_t)
\end{equation}
where $s^i_t$, $a^i_t$, $s^i_{t+1}$ are sampled from the trajectories collected from $E_{sim}$.

\subsubsection{Overall Algorithm}
In this part, we will write a clear pseudo-code in Algorithm 1 to show how the process is implemented. 

\begin{algorithm}[h!]
\DontPrintSemicolon
\caption{Algorithm for \ours}
\label{algo:UGAT}
\KwIn{Initial policy $\pi_{\theta}$, forward model $f_{\phi^+}$, inverse model $h_{\phi^-}$, domain info set $\mathcal{K}_d$, real-world dataset $\mathcal{D}_{real}$, simulation dataset $\mathcal{D}_{sim}$}
\KwOut{Policy $\pi_{\theta}$, $f_{\phi^+}$, $h_{\phi^-}$}

Pre-train policy $\pi_{\theta}$ for M iterations in $E_{sim}$ \;

\For {i = 1,2, \dots, I}
{
    Rollout policy $\pi_{\theta}$ in $E_{sim}$ and add data to $\mathcal{D}_{sim}$ \;
    Load corresponding $\mathcal{K}_d^i$ (weather, road condition)\;
    Rollout policy $\pi_{\theta}$ in $E_{real}$ and add data to $\mathcal{D}_{real}$ \\

    \# \textbf{\textit{Forward model update}} \;
    \For {l = 1, 2, \dots, n}{
        Acquire $D_{r}^l$ dynamics by Equation~\eqref{eq:llm}\;
        Feature fusion $x_t^l$ follow Equation~\eqref{eq:fuison1}, ~\eqref{eq:fuison2}\;
        Construct $\mathcal{X}_t = (x_t^1, x_t^2, \ldots, x_t^n)$\;
        Predict $\hat{s}_{t+1}$ by $f_{\phi^+}$ as Equation~\eqref{eq:forward2}
    }
    
    Update $f_{\phi^+}$ with Equation~\eqref{eq:forward-loss} \;
    \# \textbf{\textit{Inverse model update}} \;
    Update $h_{\phi^-}$ with Equation~\eqref{eq:inverse-loss}  \;
    
    \# \textbf{\textit{Policy training}}\;
    \For {e = 1, 2, \dots, E}
    {
        \# \textbf{\textit{Policy update step}}\;
        Improve policy $\pi_{\theta}$ with reinforcement learning\;
    }  
}
\end{algorithm} 

\section{Experiments}
The experimental setups and results interpretation will be elaborated in this section. The implementation of \ours will also be introduced based on a cross-simulator platform, Libsignal.

\subsection{\textbf{Experiment Settings}}
In this section, we introduce the overall environment setup for our experiments, commonly used metrics, important hyperparameters and model structures. \\

\begin{table}[htb]
\small
\centering
\caption{Real-world Configurations for $E_{real}$}
\label{tab:param}
\setlength{\tabcolsep}{1mm}

\begin{tabular}{cccccc}
\toprule
Setting & \begin{tabular}[c]{@{}c@{}}accel \\ (m/$s^2$)\end{tabular} & \begin{tabular}[c]{@{}c@{}}decel \\ (m/$s^2$)\end{tabular} & \begin{tabular}[c]{@{}c@{}}eDecel \\ (m/$s^2$)\end{tabular} & \begin{tabular}[c]{@{}c@{}}sDelay \\ (s)\end{tabular} & Description        \\ \midrule
V0 & 2.60 & 4.50 & 9.00 & 0.00 & Default setting \\
V1   &   1.00     &  2.50 & 6.00 & 0.50 & Lighter loaded vehicles   \\
V2   &   1.00     &  2.50 & 6.00 & 0.75 & Heavier loaded vehicles \\
V3   &   \textcolor{black}{0.75}     &  \textcolor{black}{3.50} & \textcolor{black}{6.00} & \textcolor{black}{0.25}    &  Rainy weather\\
V4   &   \textcolor{black}{0.50}     &  \textcolor{black}{1.50} & \textcolor{black}{2.00} & \textcolor{black}{0.50} & Snowy weather   \\
\bottomrule
\end{tabular}
\end{table}

\subsubsection{\textbf{Environment Setup}}
In our study, we leverage the LibSignal~\cite{mei2022libsignal} traffic signal control library, an open-source framework that incorporates multiple simulation environments. Our implementation involves using Cityflow \cite{zhang2019cityflow} as the simulation environment $E_{sim}$ and SUMO \cite{lopez2018microscopic} as the real-world environment $E_{real}$. Throughout the paper, we consistently refer to $E_{sim}$ and $E_{real}$ as our default simulation and real-world environments, respectively. It is important to note that this simulation-to-simulation setting not only serves as a representative sim-to-real scenario but also allows for replicable and reproducible results in our experiments.

To simulate real-world scenarios, we consider four different configurations in SUMO, representing two types of real-world scenarios: heavy industry roads and special weather-conditioned roads, with specific parameter settings detailed in Table~\ref{tab:param}.
The four configurations follows \cite{da2023uncertainty}:

\begin{table*}[h!]
\caption{The performance using \textbf{Direct-Transfer}, \textbf{Vanilla-GAT} compared with using \textbf{\ours} method. The ($\cdot$) shows the metric gap $\psi_{\Delta}$ from $E_{real}$ to $E_{sim}$ and the $\pm$ shows the standard deviation with 5 runs. The $\uparrow$ means that the higher value for the metric indicates a better performance and $\downarrow$ means that the lower value indicates a better performance. }
\label{tab:result}
\centering
\setlength{\tabcolsep}{1mm}
\scalebox{0.9}{
\begin{tabular}{ccccccc}
\toprule
Setting & Methods & \multicolumn{5}{c}{Metrics} \\ 

\cmidrule(lr){3-7}

& - &$\att(\Delta\downarrow)$ &$\tp(\Delta\uparrow)$ &$\reward(\Delta\uparrow)$ &$\queue(\Delta\downarrow)$ &$\delay(\Delta\downarrow)$   \\

\midrule
\multirow{3}{*}{$V1$} &
Direct-Transfer  & 158.93 (47.69)$_{\pm{\text{55.02}}}$ & 1901 (-77)$_{\pm{\text{52.21}}}$ & -71.55 (-32.11)$_{\pm{\text{22.51}}}$ & 47.71 (21.59)$_{\pm{\text{14.98}}}$ & 0.73 (0.11)$_{\pm{\text{0.03}}}$ \\
& Vanilla-GAT   & 156.10 (44.87)$_{\pm{\text{4.81}}}$ & 1905 (-73)$_{\pm{\text{13.00}}}$ & -70.03 (-30.59)$_{\pm{\text{3.80}}}$ & 46.61 (20.50)$_{\pm{\text{1.97}}}$ & 0.71 (0.09)$_{\pm{\text{0.01}}}$ 
\\

& \ours  & \textbf{154.97 (43.74)$_{\pm{\text{6.09}}}$} & \textbf{1918 (-60)$_{\pm{\text{9.62}}}$} & \textbf{-66.88 (-27.44)$_{\pm{\text{4.47}}}$} & \textbf{44.56 (18.45)$_{\pm{\text{2.96}}}$} & \textbf{0.71 (0.09)$_{\pm{\text{0.01}}}$}   \\

\midrule
\multirow{3}{*}{$V2$} &
Direct-Transfer  & 177.27 (66.03)$_{\pm{\text{82.63}}}$ & 1898 (-80)$_{\pm{\text{102.25}}}$ & -87.71 (-48.27)$_{\pm{\text{26.18}}}$ & 58.59 (32.47)$_{\pm{\text{17.46}}}$ & 0.76 (0.14)$_{\pm{\text{0.02}}}$ \\

& Vanilla-GAT  & 180.58 (69.35)$_{\pm{\text{11.72}}}$ & \textbf{1908 (-69)$_{\pm{\text{12.00}}}$} & -89.69 (-50.25)$_{\pm{\text{9.13}}}$ & 59.93 (33.82)$_{\pm{\text{8.01}}}$ & 0.74 (0.12)$_{\pm{\text{0.07}}}$ 
\\

& \ours  & \textbf{174.31 (63.08)$_{\pm{\text{13.11}}}$} & 1904 (-73)$_{\pm{\text{21.63}}}$ & \textbf{-84.71 (-45.27)$_{\pm{\text{18.89}}}$} & \textbf{56.64 (30.53)$_{\pm{\text{12.62}}}$} & \textbf{0.72 (0.10)$_{\pm{\text{0.02}}}$} \\

\midrule
\multirow{3}{*}{$V3$} &
Direct-Transfer  &  205.86 (94.63)$_{\pm{\text{64.49}}}$ & 1877 (-101)$_{\pm{\text{100.86}}}$ & -101.26 (-61.82)$_{\pm{\text{20.10}}}$ & 67.62 (41.51)$_{\pm{\text{13.37}}}$ & 0.76 (0.14)$_{\pm{\text{0.03}}}$ 
\\

& Vanilla-GAT  &  214.29 (103.06)$_{\pm{\text{40.59}}}$ & 1846 (-131)$_{\pm{\text{56.74}}}$ & -91.15 (-51.71)$_{\pm{\text{15.13}}}$ & 60.93 (34.82)$_{\pm{\text{10.10}}}$ & 0.73 (0.11)$_{\pm{\text{0.02}}}$ \\

& \ours  &  \textbf{198.48 (87.25)$_{\pm{\text{7.27}}}$} & \textbf{1879 (-98)$_{\pm{\text{6.02}}}$} & \textbf{-89.25 (-49.81)$_{\pm{\text{5.51}}}$} & \textbf{59.65 (33.54)$_{\pm{\text{3.70}}}$} & \textbf{0.72(0.10)$_{\pm{\text{0.01}}}$} \\

\midrule
\multirow{3}{*}{$V4$} &
Direct-Transfer  & 332.48 (221.25))$_{\pm{\text{109.00}}}$ & 1735 (-252)$_{\pm{\text{151.91}}}$ & -126.71 (-87.23)$_{\pm{\text{14.79}}}$ & 84.53 (58.42)$_{\pm{\text{9.86}}}$ & 0.83 (0.21)$_{\pm{\text{0.01}}}$ \\

& Vanilla-GAT  & 318.70 (207.47)$_{\pm{\text{12.35}}}$ & 1750 (-227)$_{\pm{\text{16.93}}}$ & -115.01 (-75.57)$_{\pm{\text{9.27}}}$ & 76.74 (50.63)$_{\pm{\text{5.10}}}$ & 0.81 (0.19)$_{\pm{\text{0.08}}}$  \\

& \ours  & \textbf{310.29 (199.06)$_{\pm{\text{22.57}}}$} & \textbf{1750 (-227)$_{\pm{\text{16.47}}}$} & \textbf{-113.55 (-74.11)$_{\pm{\text{6.68}}}$} & \textbf{75.77 (49.66)$_{\pm{\text{4.48}}}$} & \textbf{0.81 (0.19)$_{\pm{\text{0.01}}}$}  \\

\bottomrule
\end{tabular}
}

\label{tab:main}

\end{table*}

\noindent$\bullet$ {\emph{V0: Default setting\footnote{\url{https://sumo.dlr.de/docs/Definition_of_Vehicles,_Vehicle_Types,_and_Routes.html}}}}. Default parameters for SUMO and CityFlow, profiling the ideal settings in $E_{sim}$.
\\
$\bullet$ {\emph{V1 \& V2: Heavy industry roads.}}
In this configuration, we model the situation that the commonly visited vehicles are of different types. $V1$ describes roads with lighter-loaded vehicles, while $V2$ represents the same roads with heavier-loaded vehicles, they also differ in startup delay.
\\
$\bullet$ {\emph{V3 \& V4: Special weather-conditioned roads.}}
Consider areas with special weather conditions, $V3$ and $V4$ represent rainy and snowy weather conditions, respectively.

\begin{table}[htb]
    \centering
    \caption{Overall performance in $E_{sim}$}
    \scalebox{0.82}{
    \begin{tabular}{cccccc}
        \toprule
        $Env$ & $ATT$ & $TP$ & $Reward$ & $Queue$ & $Delay$\\ \midrule
        \textbf{$E_{sim}$} & \textbf{111.23$_{\pm{\text{3.5}}}$} & \textbf{1978$_{\pm{\text{5}}}$} &\textbf{-39.44$_{\pm{\text{2.23}}}$} &\textbf{26.11$_{\pm{\text{1.15}}}$} &\textbf{0.62$_{\pm{\text{0.10}}}$} \\
        \bottomrule
    \end{tabular}}
    \label{tab:simresult}
\end{table}

\subsubsection{\textbf{Evaluation Metrics}} \label{title:metrics}
The primary goal of this work is to mitigate the performance gap of the trained policy $\pi_{\theta}$ in the simulation environment $E_{sim}$ and in the real-world environment $E_{real}$. 
We calculate the performance difference $\Delta$ for commonly used traffic signal control metrics: average travel time, throughput, reward, average queue length, and average delay, and their detailed definition can be found in \cite{da2023uncertainty}. We denote their differences as $\att_{\Delta}$, $\tp_{\Delta}$, $\reward_{\Delta}$, $\queue_{\Delta}$, and $\delay_{\Delta}$. For a given metric $\psi$:
\begin{equation} \label{eq:delta}
\psi_{\Delta} = \psi_{real} - \psi_{sim}
\end{equation}
Since in real-world settings, policy $\pi_{\theta}$ tends to perform worse than in simulation, the values of $\att$, $\queue$, and $\delay$ in $E_{real}$ are typically larger than those in $E_{sim}$. Based on our goal of mitigating the gap and improving the performance of $\pi_{\theta}$ in $E_{sim}$, we expect that for $\att_{\Delta}$, $\queue_{\Delta}$, and $\delay_{\Delta}$, smaller values are better, while for $\tp_{\Delta}$ and $\reward_{\Delta}$, larger values are better. For a fair comparison, $\psi_{sim}$ of all methods in $E_{sim}$ are trained to be similar and reported in Table~\ref{tab:simresult}: with the similar $\psi_{sim}$, we can also compare $\psi_{real}$ from different methods to know which method performs the best.

\subsection{Experimental Results and Analysis}

We analyze the proposed method in the following way: First, we verify if the prompt result from Large Language Model is giving the rational inference based on the context description. Second, we compare to the Direct-Transfer to verify if \ours is able to mitigate the performance. In part 3, we discuss the performance improvement competing with baseline models. Furthermore, we demonstrate our method's contribution to the forward model's accuracy and how it is correlated to the final $E_{real}$ performance gap mitigation.

\paragraph{Prompt Intention Analysis}
In this section, we conduct an analysis on the verification of whether the LLM Prompt infers the expected information following the practical laws in realistic. We conduct prompts in the format of $\langle \textit{Task} \rangle \langle [\textit{Context}] \rangle \langle \textit{Output Restriction}\rangle$ as defined in Equation~\eqref{eq:define}. We first introduce the $\langle \textit{Task} \rangle$ description below:
\begin{tcolorbox}[colback=green!5,
                  colframe=black,
                  width=8.5cm,
                  arc=1mm, auto outer arc,
                  boxrule=0.05pt,
                  fontupper=\small, 
                 ]
The indicators describing the traffic dynamics include the average acceleration (AC) of the vehicles (m/s²), the average deceleration (AD) (m/s²), the average emergency deceleration (AED) (m/s²) and the average startup delay (ADL) describing the average time needed for the waiting vehicles to start moving with the unit (s), and the above might vary based on weather or road type. Please assume the above indicators based on the traffic perceptive information below:
\end{tcolorbox}
Then specifically, for the $\langle[\textit{Context}]\rangle $, we have the following implementation (the colored content is replaceable based on the actual situation, \weather{weather}, \road{road}, and \traffic{traffic state (vehicle quantity)} in correspondence with Equation~\eqref{eq:promptpair}:
\begin{tcolorbox}[colback=gray!10,
                  colframe=black,
                  width=8.5cm,
                  arc=1mm, auto outer arc,
                  boxrule=0.05pt,
                  fontupper=\small, 
                 ]
V1: In \weather{sunny} day, on a \road{light industry road} with \traffic{8} vehicles, \\
V2: In \weather{sunny} day, on a \road{heavy industry truck road}, \traffic{5} vehicles, \\
V3: In \weather{rainy} day, on a \road{normal road} with \traffic{10} vehicles, \\
V4: In \weather{snowy} day, on a \road{normal road} with \traffic{7} vehicles. 
\end{tcolorbox}
And for $\langle \textit{Output Restriction}\rangle$ we design as below: 
\begin{tcolorbox}[colback=purple!5,
                  colframe=black,
                  width=8.5cm,
                  arc=1mm, auto outer arc,
                  boxrule=0.05pt,
                  fontupper=\small, 
                 ]
Please answer by replacing \{value\} in the format below: \par
[average acceleration: {value}], \par
[average deceleration: {value}], \par
[average emergency deceleration: {value}], \par
[average startup delay: {value}].
\end{tcolorbox}

\begin{figure}[htb]
    \centering
    \includegraphics[width=0.47\textwidth]{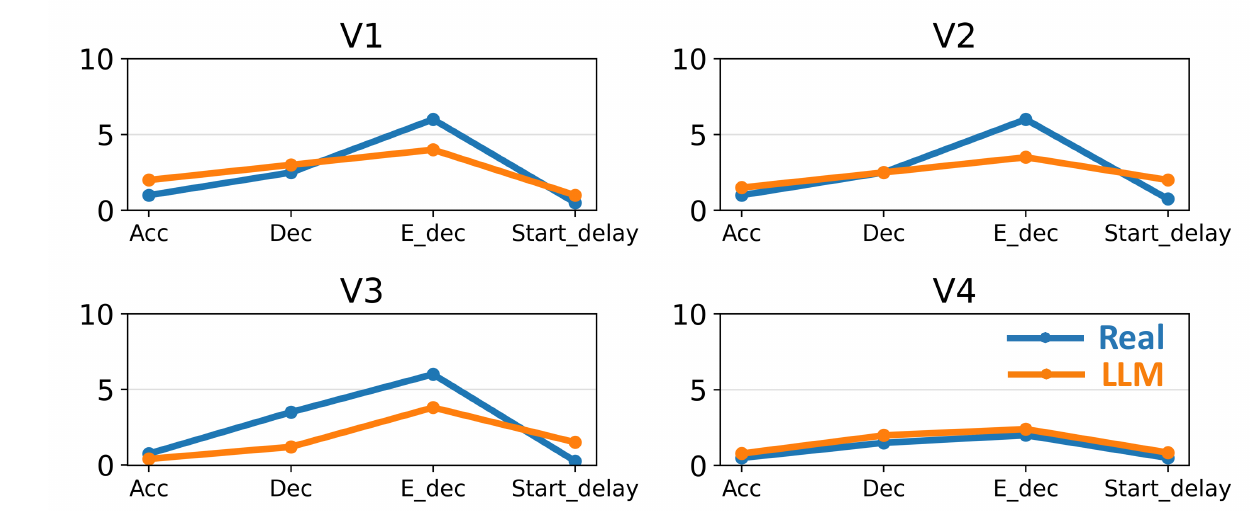}
    \vspace{-3mm}
    \caption{Comparison of LLM prompt answers and real-world settings reflects the same tendency across 4 versions.}
    \label{fig:compare_prompt}
\end{figure}

We take the real-world setting dynamics values as the ground truth and compare the LLM inferred value outputs to the ground truth to analyze their relationship, the results are shown in Figure~\ref{fig:compare_prompt}. We could observe that within each sub-figure, LLM's inference results show a similar curve across metrics, and from v1 to v4, the LLM is also reflecting the same tendency as shown by Real settings. This proves the LLM's ability to provide a realistic inference based on the given information, thus guaranteeing the rationality to apply Prompt in our task of approximating real-world dynamics.

\begin{figure}[t!]
    \centering
    \begin{tabular}{cc}
    \includegraphics[width=0.220\textwidth]     
    {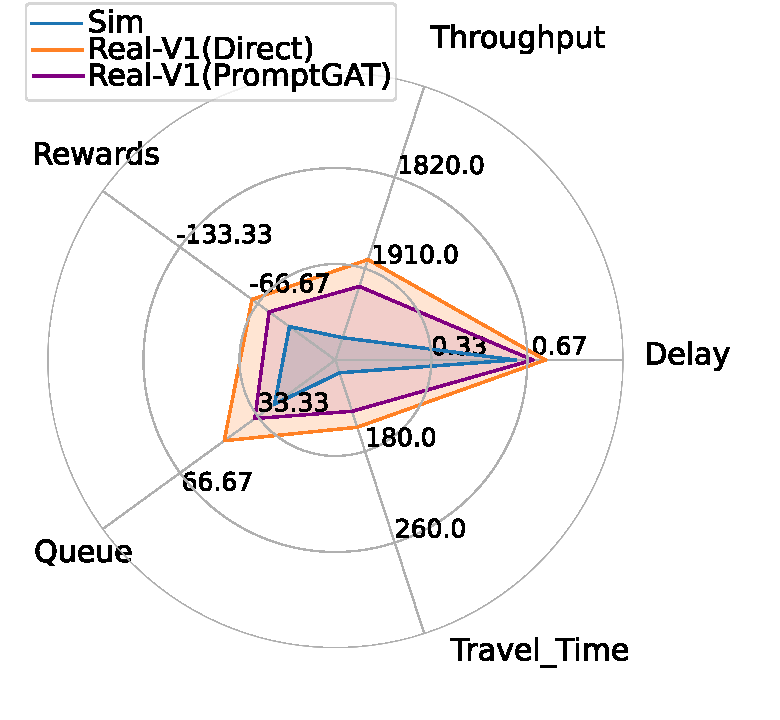}
    &\includegraphics[width=0.220\textwidth]
    {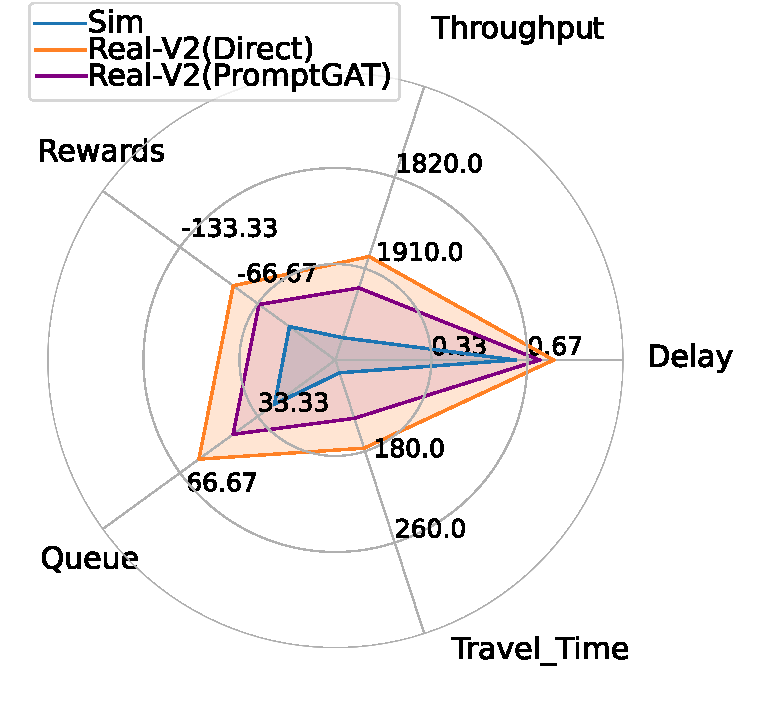}\\
    \includegraphics[width=0.220\textwidth]     
    {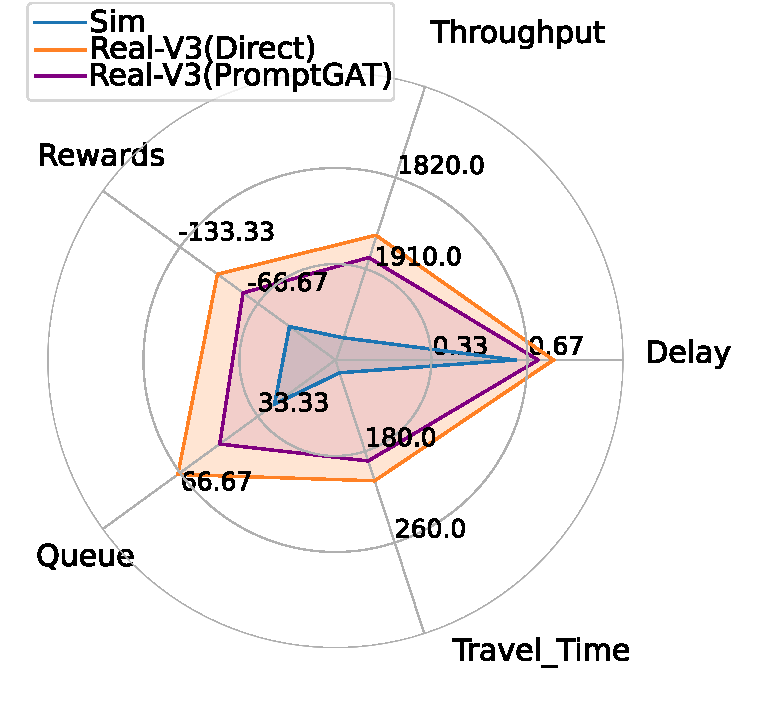}
    &\includegraphics[width=0.220\textwidth]
    {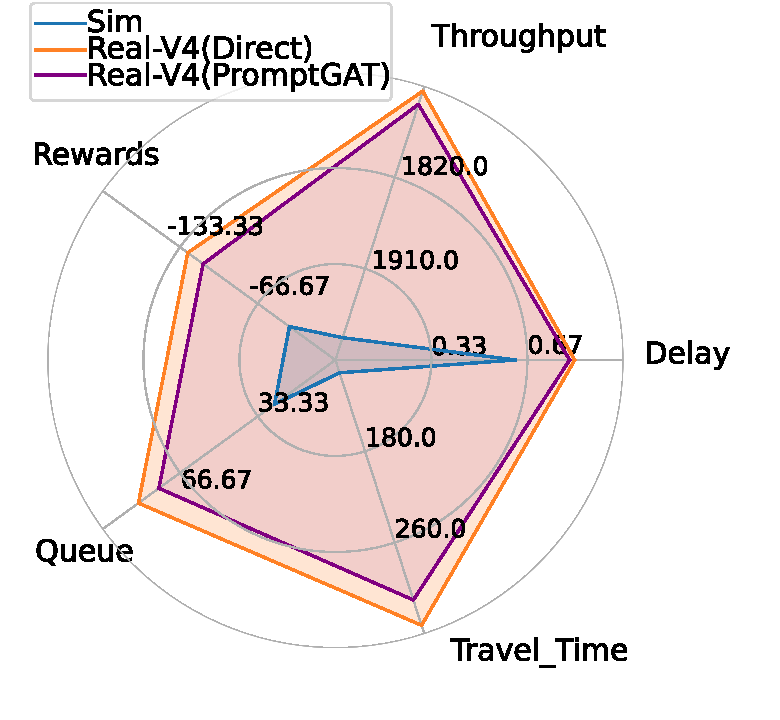}\\

    \end{tabular}
    
    \vspace{-3mm}
    \caption{The performance in the $E_{real}$ using Direct-Transfer and \ours comparing to the performance in $E_{sim}$.}
    \label{fig:shrinkgap}
\end{figure}

\paragraph{Ability to Mitigate the Performance Gap}
We first train policies using DQN in $E_{sim}$ to well-converged as in Table~\ref{tab:simresult} and apply to 4 $E_{real}$ settings described in Table~\ref{tab:param}. This is taken as the `direct transfer', which exists a large gap compared to $E_{sim}$ training performance~\cite{da2023uncertainty}. Then we leverage the proposed method \ours to train policies under the same 4 settings. Following the metrics in Section~\ref{title:metrics}, we could show a comparison in Figure~\ref{fig:shrinkgap}: the center blue area is the metrics connection reported when policies are well trained in $E_{sim}$, these are the most ideal achievement that a policy could acquire. When the well-trained policy directly applies to $E_{real}$, the performance is shown in the orange area, we could obviously observe that large gaps commonly exist in 4 different real-world settings. Even though the severeness varies on settings, still non of the gap is trivial. \ours shows promising results by effectively shrinking the gap to a much lower level (as shown in the purple area).

\paragraph{Comparison to Baseline Models}
In this part, we analyze how the proposed \ours compete with other methods at a quantity level, including Direct Transfer and \vanilla. We apply all 3 approaches in 4 settings and compare their performance under 5 metrics. Each performance is represented as mean value and standard deviation after conducting 5 runs of tests. As shown in Table~\ref{tab:result} that most of the time, the \ours performs better than other baselines across various settings and metrics.

\subsection{Further Study and Experiment Details}\label{casestudy}
In this section, we conduct a case study on setting $V4$  to show contribution of \ours to the forward model accuracy and its relation to the performance gap mitigation in $E_{real}$.

\paragraph{Correlation Analysis}

\begin{figure}[h!]
    \centering
    \includegraphics[width=0.5\textwidth]{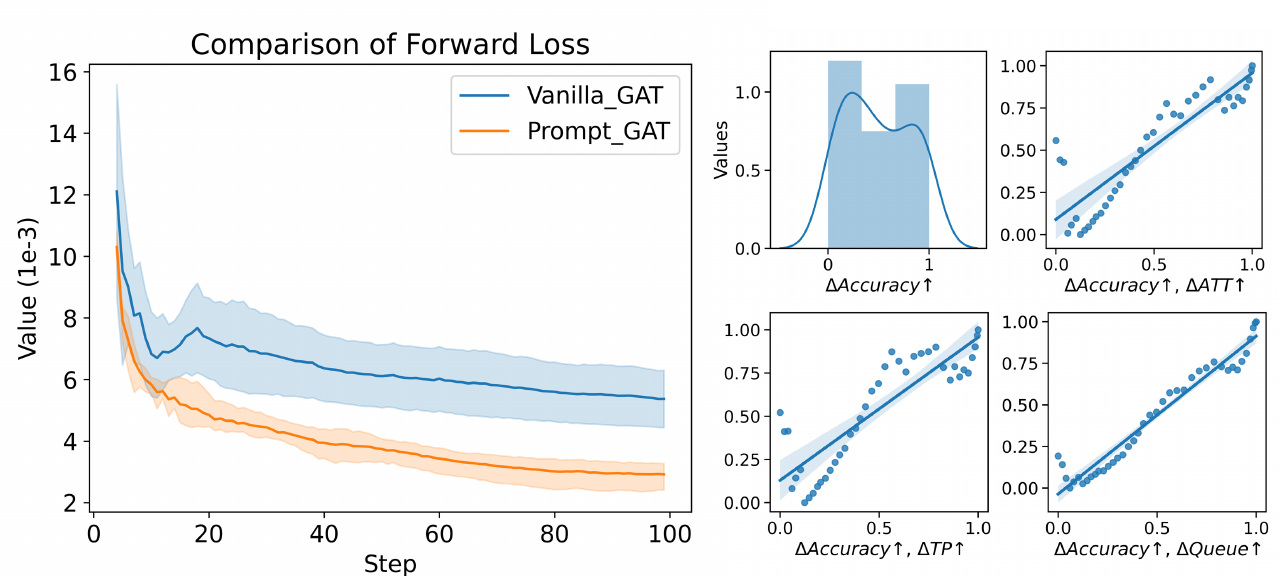}
    \vspace{-3mm}
    \caption{Left: Prediction error from the forward model using \ours $\textit{vs}$ VanillaGAT, our method consistently approximates the true system dynamics and reduces the loss. Right: The correlation between improvement of accuracy and improvement of mitigated performance gap $\Delta$ in $E_{real}$.}
    \vspace{-3mm}
    \label{fig:forward}
\end{figure}

\begin{figure}[h!]
    \centering
    \includegraphics[width=0.5\textwidth]{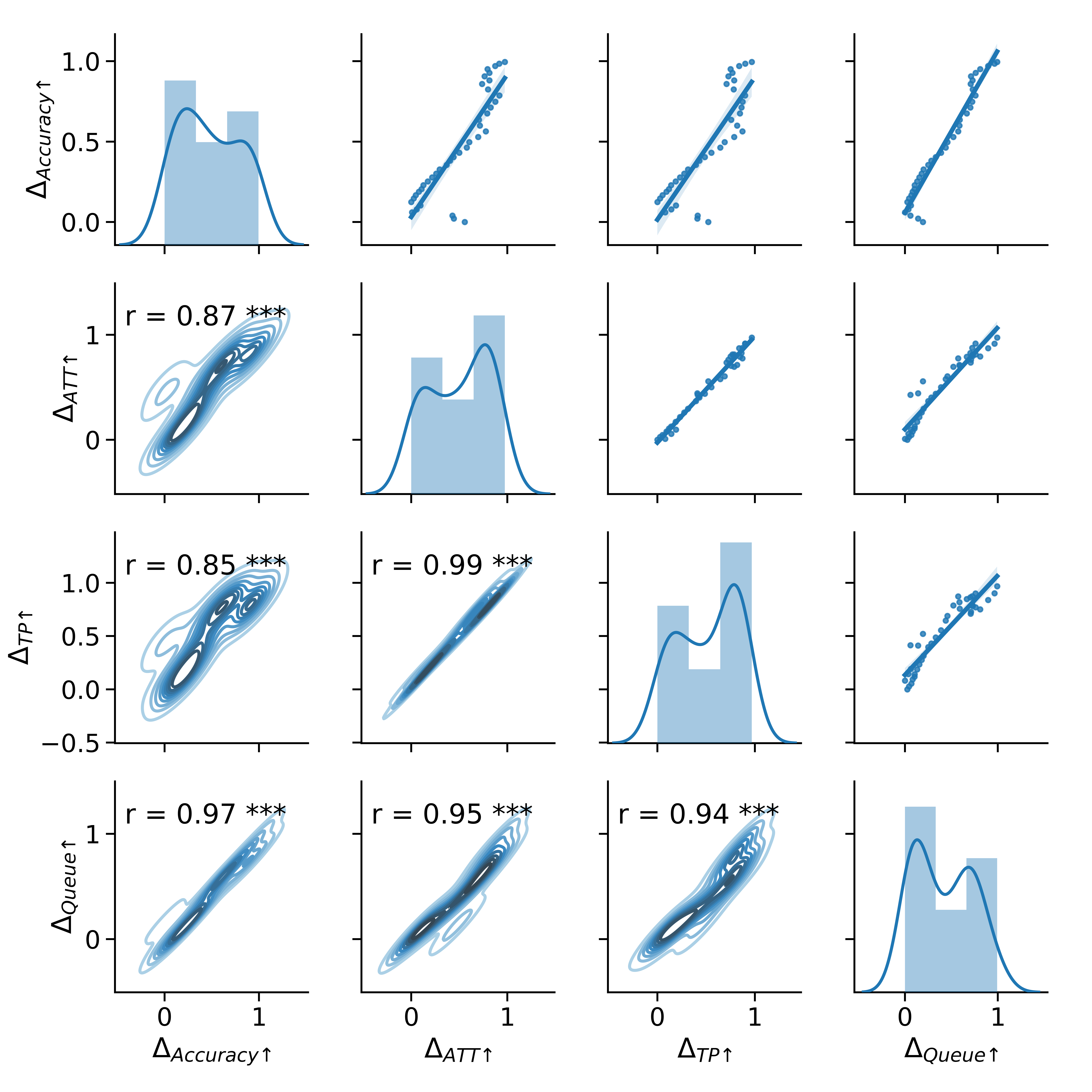}
    \caption{Correlation analysis between improvements of forward model accuracy (compared with Vanilla-GAT) and the improvements of
performance in $E_{real}$. The value of $r$
indicates the Pearson Correlation Coefficient, and the values
with $\ast$ indicating statistical significance for correlation, where
$\ast\ast\ast$ indicates the p-value for testing non-correlation $p \leq$  0.001.}
    \label{fig:correlation}
\end{figure}

We first compare the prediction error of our method to Vanilla GAT as in Figure~\ref{fig:forward} (left). Proving our method provides a better prediction of system dynamics. In order to understand how would dynamic profiling ability influence the model's performance gap in $E_{real}$, we conduct correlation analysis across the metrics gap of average travel time, throughput, and waiting queue length. As shown in Figure~\ref{fig:forward} (right), the improvement of the mitigated gap (absolute values) across multiple metrics is positively correlated to the improvement of the forward model's accuracy. This indicates \ours mitigates the real-world gap by better profiling the realistic system dynamics. A more detailed analysis including  p-value testing as shown in Figure~\ref{fig:correlation}.

The values of $\Delta_{\text{Accuracy↑}}$, $\Delta_{\text{ATT↑}}$, $\Delta_{\text{TP↑}}$, $\Delta_{\text{Queue↑}}$ indicating how much gap is mitigated, they are further definitions based on Equation~\eqref{eq:delta}:
\begin{equation}\label{eq:gapimprove}
    \Delta_{\text{↑}} = \lvert \psi_{a\Delta} - \psi_{b\Delta} \rvert
\end{equation}

where $a$ and $b$ are two approaches used, here they are Vanilla-GAT and \ours, respectively. Based on the absolute value of this equation, since the \ours performs consistently better than Vanillia-GAT from Table~\ref{tab:result}, the larger value $\Delta_{\text{↑}} $ is, the much more gap is mitigated. And since every metric has its own range, for unity, we normalized the $\Delta_{\text{↑}}$ using max-min normalization. Apart from the commonly used metrics, the $\Delta_{\text{Accuracy↑}}$ refers to how much accuracy (loss) has been improved. 

We could observe from Figure~\ref{fig:correlation} that for the relation between the accuracy improvement $\Delta_{\text{Accuracy↑}}$ and each of the metrics, they are strongly correlated except for minor outlier points, which proves that, when the \ours provides a better depiction inference on the system dynamics, the inverse model's action would be better grounded to the realistic scenario, policy $\pi$ will learn in a more robust way and leading to a final lower performance gap in reality. Other metrics' mutual comparison also reflects the correctness of our method by showing a strong correlation between each other.

\section{Conclusion}

In this paper, we propose a prompt-based grounded action transformation method named \ours in the traffic signal control domain to mitigate the sim-to-real performance gap. By leveraging the inference ability from pre-trained large language models, incorporating the perceptible domain knowledge, \ours manages to better profile and depict the system dynamics in real-world settings and successfully increase the forward model's prediction accuracy in the GAT process. Benefiting from the more accurate real-world dynamics prediction, the grounded action transformation is taken to enable the agent to perform well in the real world, further improving the model's transferring performance. Sufficient experiments demonstrate the rationale of LLM's inference process and prove \ours has the potential to solve the police's real-world deployment challenge. This research first leverages the LLMs to solve policies' real-world deployment problems, cast hope for the future exploration.
\\

\bibliography{aaai24}


\end{document}